\documentclass[letterpaper, 10 pt, conference]{ieeeconf}  

\IEEEoverridecommandlockouts                              

\usepackage{balance}
\usepackage{color}

\usepackage{graphics} 
\usepackage{amsmath,amssymb,amsfonts} 
\usepackage{bm}
\usepackage{subcaption}	 
\usepackage{graphicx}
\usepackage{multirow}
\usepackage{array}

\usepackage{url}
\usepackage{hyperref}
\usepackage{xcolor}
\usepackage{}

\usepackage{float}
\usepackage{booktabs}
\usepackage{multirow}
\usepackage{textcomp}
\usepackage{setspace}
\usepackage{tabularx}
\usepackage{makecell}
\usepackage{colortbl}  
\usepackage{pifont}

\newcommand{\PAR}[1]{\vskip4pt \noindent{\bf #1~}}

\usepackage[backend=bibtex,natbib=true,style=numeric-comp,sorting=none,giveninits=true,maxbibnames=99,url=false,doi=false]{biblatex}
\usepackage{censor}
\title{\LARGE \bf
SparseWorld: Enhancing End-to-End Autonomous Driving via World Models with Sparse Scene Representation
}

\author{
Ruoyu Wang$^{1,*}$, Jingke Wang$^{1,*}$, Yukai Ma$^{1,\dagger}$, Yuehao Huang$^{1}$, Shuangming Lei$^{1}$, \\
Guanglin Xu$^{2}$,  Aixue Ye$^{2}$, Yong Liu$^{1,\ddagger}$
\thanks{This work was supported by "Zhejiang Key Laboratory of Advanced Intelligent Warehousing and Logistics Equipment" (Grant No. 2024E10007).}%
\thanks{$^{1}$ The authors are with the Institute of Cyber-Systems and Control, Zhejiang University, Hangzhou, China.}%
\thanks{$^{2}$ The authors are with the 2012 Labs, Huawei.}%
\thanks{$^{*}$ These authors contributed equally to this work.}%
\thanks{$^{\dagger}$ The project leader of this work.}%
\thanks{$^{\ddagger}$ Corresponding author (Email: {\tt\small yongliu@iipc.zju.edu.cn}).}%
}

\begin{document}

\maketitle
\thispagestyle{empty}
\pagestyle{empty}

\begin{abstract}
Recently, world models have made significant progress in enhancing end-to-end driving systems through both future situation forecasting and improved scene understanding. 
However, existing driving world models are typically built upon dense scene representations, causing high computational costs and redundant information.
In this paper, we present SparseWorld, a lightweight world model that focuses on predicting only the critical layout of the scene, enabling efficient future forecasting for end-to-end driving systems. 
SparseWorld first performs autoregressive rollout to forecast future map elements and surrounding agents, enabling the model to learn how driving scenarios evolve over time. It then leverages these predicted futures to refine downstream motion prediction and trajectory planning.
Specifically, we propose a Sparse Dreamer that anticipates future instances in the latent space through joint temporal and spatial attention. 
By interacting with predicted future instances, the motion planner captures more accurate motion patterns and generates more informed and safety-aware trajectories.
Extensive experiments demonstrate that SparseWorld significantly reduces collision risk and achieves state-of-the-art performance on the open-loop planning metrics of the nuScenes dataset with a collision rate of 0.05\%. 
Moreover, it substantially outperforms the baseline method in closed-loop planning metrics on the Bench2Drive benchmark.
Supplementary material is available at the project page: \url{https://wryzju.github.io/SparseWorld/}.
\end{abstract}
\section{Introduction}
\label{sec:intro}



\begin{figure}[t]
      \centering
      \includegraphics[width=\linewidth]{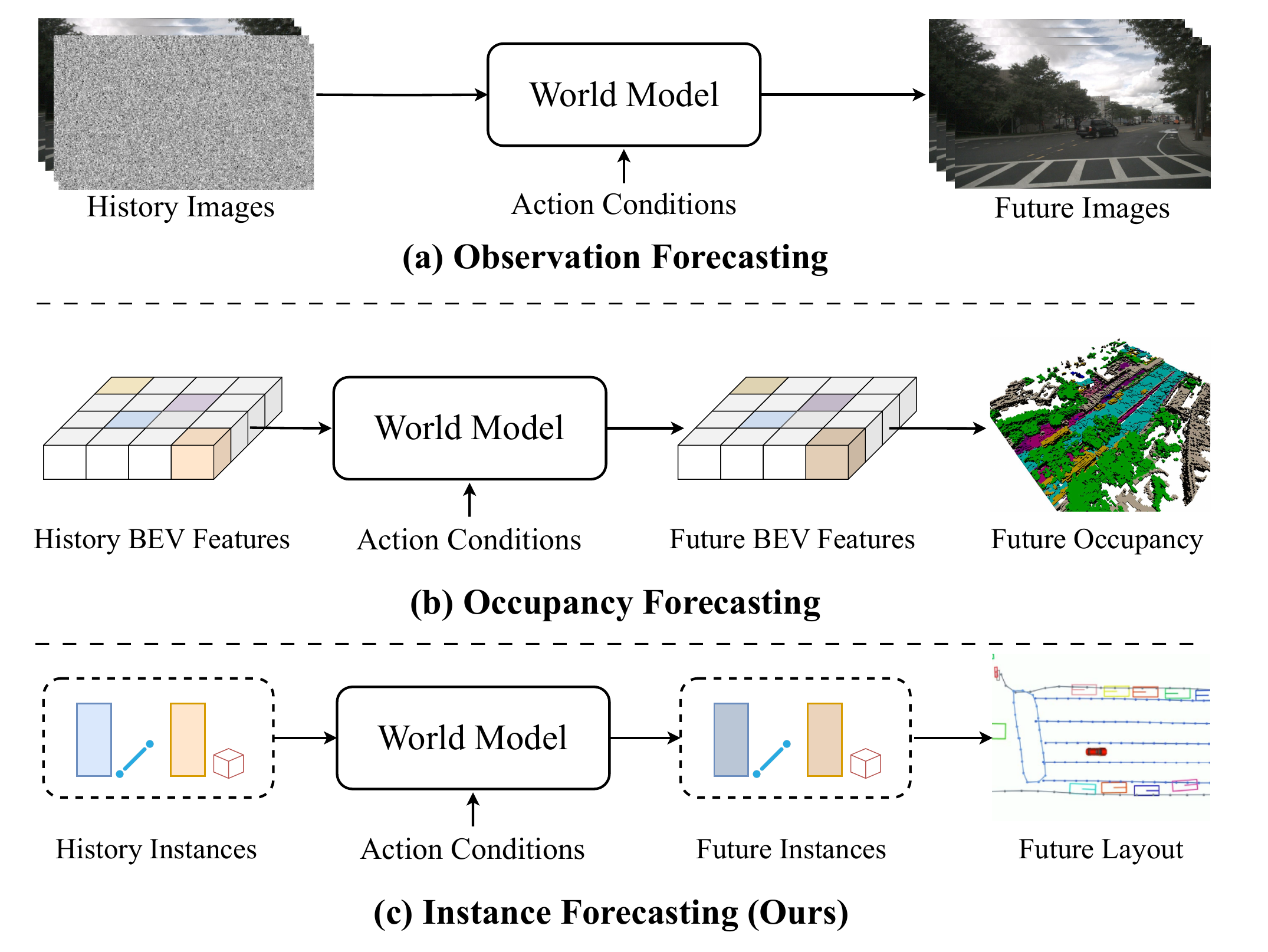}
      \captionsetup{font=small}
      \caption{Comparison of different designs of driving world model frameworks.
        (a) The world model is first introduced in autonomous driving to generate future observations conditioned on actions.
        (b) The BEV latent world model is proposed to better preserve the 3D semantic structure of the scene.
        (c) Our lightweight design generates only agents and map layouts, improving efficiency while enhancing end-to-end planning.
        }
      \label{fig:teaser}
      \vspace{-20pt} 
\end{figure}

Modeling the evolution of driving scenes with world models can greatly enhance the safety and accuracy of trajectory planning in end-to-end autonomous driving. As shown in Fig.~\ref{fig:teaser}, recent research in this area has followed two primary directions. The first approach focuses on observation forecasting. For example, models like DriveWM~\cite{wang2024driving} and VISTA~\cite{gao2024vista} generate future visual observations to simulate upcoming scenarios. The second approach forecasts scene evolution using dense BEV (bird’s-eye view) latent representations. For instance, Drive-OccWorld~\cite{yang2025driving} employs an autoregressive framework to predict future occupancy maps. In a different vein, methods like WOTE~\cite{li2025end} bypass explicit scene prediction, instead leveraging latent future BEV features to directly inform the planning process.
Despite their promise, above approaches rely on dense future prediction, which introduces substantial computational overhead and hinders their deployment in real-world applications. This underscores the urgent need for an efficient, lightweight world model that enhances the practicality and scalability of driving systems.

World models are inspired by the human ability to reason about the dynamics of the surrounding environment. In real-world driving, human drivers naturally focus only on the future states of critical elements, such as key traffic participants and road structures, while ignoring the appearance.
This observation suggests that, for world models in autonomous driving, focusing on the future states of only salient instances in the scene is a more practical, efficient, and effective alternative to generating dense representations of the entire scene. Yet, despite its intuitive appeal and alignment with human cognition, this instance-centric approach remains largely unexplored.
In this paper, we introduce a lightweight world model focusing on predicting the critical layout of the scene via sparse scene representations.
Designed for flexibility, it can be seamlessly integrated as a plug-in module into end-to-end autonomous driving frameworks that incorporate instance-level detection.
By jointly forecasting the future states of critical traffic participants and the anticipated evolution of map elements, the proposed model significantly enhances both motion prediction and trajectory planning, thereby promoting more reliable autonomous driving.

Specifically, we propose SparseWorld, a vision-centric, lightweight world model designed for instance-level future forecasting and motion planning refinement. SparseWorld has two core functionalities:
\textbf{(1) Sparse Forecasting of Future Scenario.}
We introduce Sparse Dreamer, a key component that predicts future scenarios from historical instances and action conditions. By integrating this module with the instance memory queue and motion planner, SparseWorld generates future scenarios in sparse representation and future trajectories in an autoregressive manner.
\textbf{(2) Motion Planning Refinement with Future Scenarios.}
SparseWorld refines motion planning by conditioning the prediction on the high-dimensional features—which provide comprehensive context beyond simple geometric representation—of forecasted future instances.
This enables the motion predictor to learn more accurate and far-sighted motion patterns, while also allowing the trajectory planner to learn safer and more reliable trajectories.
Furthermore, we propose an Adaptive Trajectory Selection module to evaluate and select the safest trajectory among candidate trajectories from various origins.

We evaluate SparseWorld using the nuScenes~\cite{caesar2020nuscenes} and Bench2Drive~\cite{jia2024bench2drive} dataset. 
Empirical evaluations show that on both open-loop and closed-loop planning benchmarks, our proposed SparseWorld delivers high-fidelity forecasting and statistically significant improvements over baseline methods, SparseDrive~\cite{sun2024sparsedrive} and VAD~\cite{jiang2023vad}.
When integrated with SparseDrive, SparseWorld achieves a state-of-the-art (SOTA) open-loop collision rate of 0.05\%, representing a 37.5\% reduction relative to the baseline method. In the closed-loop setting, it attains a driving score of 48.95, reflecting a 9.9\% improvement.
Furthermore, SparseWorld substantially reduces the missing rate and demonstrates enhanced performance on the end-to-end prediction accuracy metric in the motion prediction task.
%
The primary contributions of this work can be summarized as follows: 
\begin{itemize}
\item We introduce SparseWorld, a lightweight, vision-centric world model designed for forecasting sparse future instances and refining motion planning.
\item We propose Motion Planning Refinement to enhance end-to-end autonomous driving. It utilizes future predictions to refine motion prediction and planning trajectory. A subsequent Adaptive Trajectory Selection module then selects the final trajectory from various origins to ensure safe and proactive decision-making.
\item Experimental results demonstrate that our method yields significant improvements over the baseline method in both motion prediction and open/closed-loop planning.
\end{itemize}
\section{Related Works}
\label{sec:related}
\begin{figure*}[t]
      \centering
      \includegraphics[width=\linewidth]{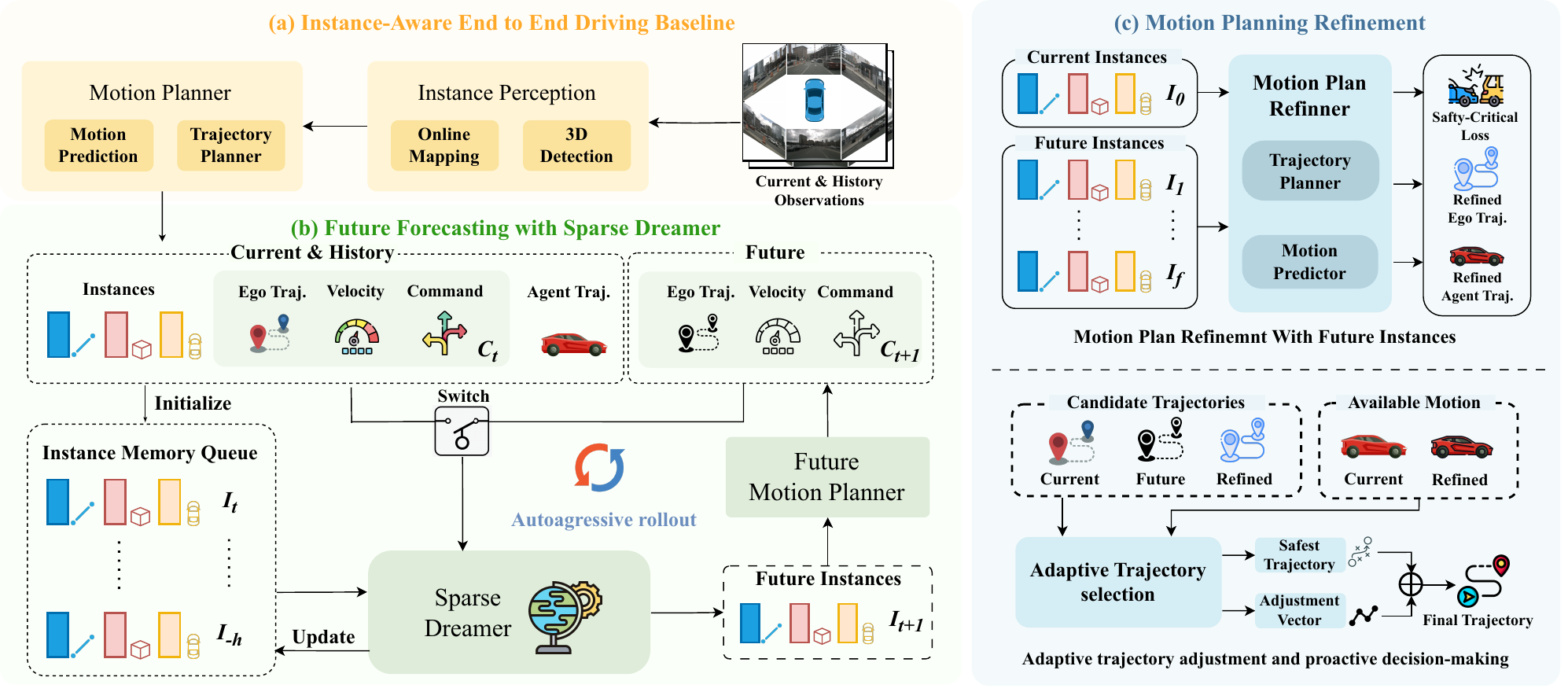}
      \vspace{-10pt}
      \captionsetup{font={small}}
      \caption{\textbf{The SparseWorld framework operates in three sequential stages.}
      (a) Instance-Aware End-to-End Driving Baseline: An instance-aware baseline model processes historical sensor data to infer agent and map instances, action conditions, and initial trajectories for both the ego vehicle and surrounding agents.
      (b) Future Forecasting with Sparse Dreamer: The Sparse Dreamer component then autoregressively generates future instances based on the historical instances and action conditions.
      (c) Motion Planning Refinement: The predicted future instances are used to refine the initial trajectories, then an adaptive trajectory selection module evaluates and chooses the safest trajectory from a set of candidates.
      }
      \label{fig:framework}
      \vspace{-20pt} 
\end{figure*}

\subsection{End-to-end Autonomous Driving}
Recent end-to-end autonomous driving studies~\cite{gao2025rad,li2025hydra,liao2025diffusiondrive,sima2025centaur,yuan2024drama} learn a unified mapping that converts raw multimodal sensor observations directly into executable trajectories or control commands. Within this framework, early perception-free methods~\cite{toromanoff2020end,wu2022trajectory,chen2020learning} primarily combined reinforcement learning with imitation learning. As the field has advanced, end-to-end systems that embed multi-task perceptual constraints have become the dominant paradigm. For instance, UniAD~\cite{hu2023planning} enhances system robustness by explicitly integrating perception, prediction, and planning, while introducing a goal-conditioned planning mechanism. Building on VAD~\cite{jiang2023vad}, VADv2~\cite{chen2024vadv2} adopts a probabilistic-planning paradigm to address the intrinsic uncertainty of driving decisions. SparseDrive~\cite{sun2024sparsedrive} further optimizes path generation through a multimodal trajectory-prediction framework, while TTOG~\cite{liu2025two} unifies motion prediction and planning in a two-stage trajectory-generation architecture. The core idea behind these methods is to leverage historical or current scene information to better model dynamic evolution, leading to trajectories with improved temporal consistency and safety.

\subsection{World Models for Autonomous Driving}
World models in autonomous driving fall into two main streams. The first focuses on image-domain generators, driven by latent-diffusion or autoregressive mechanisms~\cite{hu2023gaia,hu2022model,jia2023adriver,li2024drivingdiffusion,gao2024vista}, which aim to synthesize  multi-view future scenes directly in pixel space. DriveDreamer~\cite{wang2024drivedreamer} was the first to apply latent diffusion to multi-view driving video generation, showing that joint action-and-text conditioning can convey driving intent. DriveWM~\cite{wang2024driving} later extended this approach with unified multi-view prediction and an open-loop planning interface. The second stream comprises three-dimensional occupancy generators that rely on discretized voxels or continuous fields~\cite{zhang2023copilot4d,zheng2024occworld,min2023uniworld} to ensure spatio-temporal coherence of geometry and motion. OccSora~\cite{wang2024occsora} and DriveWorld~\cite{min2024driveworld} advanced this line of work by using 4D occupancy diffusion to produce semantically consistent and topologically coherent voxel grids over multi-second horizons. GaussianWorld~\cite{zuo2025gaussianworld} models 4D scene evolution from observations using Gaussian representations. 
UniScene~\cite{li2025uniscene} in contrast, employs a hierarchical generation process: BEV layouts first produce semantic occupancies, which are then converted into multimodal outputs for controllable synthesis via Gaussian rendering and sparse sampling.
To more efficiently exploit future scene dynamics, we propose a world model that forecasts agents and map elements, feeding these predictions back to jointly refine motion prediction and trajectory planning. 

\section{Methodology}
\label{sec:method}


\subsection{Overview}
Fig.~\ref{fig:framework} shows an overview of our approach. SparseWorld is integrated with instance-aware end-to-end driving baselines~\cite{sun2024sparsedrive, jiang2023vad} with two primary objectives: 
first, to efficiently forecast sparse representations of future instances, and second, to leverage these forecasts to improve the baseline's motion planning performance. Our framework consists of three processes: (1) Driving Baseline: This initial stage infers current instances, action conditions and trajectories from a sequence of historical sensor observations. (2) Future Forecasting with Sparse Dreamer (Sec.~\ref{sec:Future_forecasting}): In this stage, the Sparse Dreamer component autoregressively predicts future instances. (3) Motion Planning Refinement (Sec.~\ref{sec:Motion_and_Planning_Refine}): Finally, the predicted future instances are used to refine both the motion prediction and trajectory planning tasks.

\subsection{Future Forecasting with Sparse Dreamer}
\vspace{-2mm}
\label{sec:Future_forecasting}
\PAR{Future Forecasting.}
As depicted in Fig.~\ref{fig:framework}, the autoregressive future prediction process of SparseWorld encompasses three core components: 
(1) Instance Memory Queue (IMQ) maintains a cache repository of both historical and predicted future instances.
(2) Sparse Dreamer (SPD) leverages instances from the IMQ and action conditions to generate the subsequent frame's instances.
(3) Future Motion Planner (FMP) adopts an identical network architecture to the motion planner of the driving baseline, and infers future action conditions from predicted future instances.
The autoregressive future forecasting process $\mathcal{W}$ of SparseWorld can be defined as:
\begin{equation}
\mathcal{W}\left( \{I_{-h},\dots,I_0\}, \{C_{-h},\dots,C_0\} \right) = \{I_1,\dots,I_f\}
\end{equation}
Where the tuple $I_t =\{(\mathcal{A}_t, \mathcal{F}_t) \,|\, t \in \{-h, -h+1, \cdots, f\}\}$ denotes instances. Here, $\mathcal{A} = \{a^\mathrm{ego}, a^\mathrm{agent}, a^\mathrm{map}\}$ contains the geometric information of the anchors for the ego vehicle, surrounding agents, and map elements, 
while $\mathcal{F}$ 
holds their corresponding features. We consider a historical sequence of action conditions over $h$ timesteps, represented as $C_t = \{(v_t, \tau_t, s_t) \,|\, t \in \{-h, -h+1, \cdots, 0\}\}$. Each action condition consists of the ego velocity $v_t$, planned trajectory $\tau_t$, and steering command $s_t$. Our goal is to predict future sequences over a prediction horizon of length $f$.

\begin{figure}[t]
      \centering
      \includegraphics[width=\linewidth]{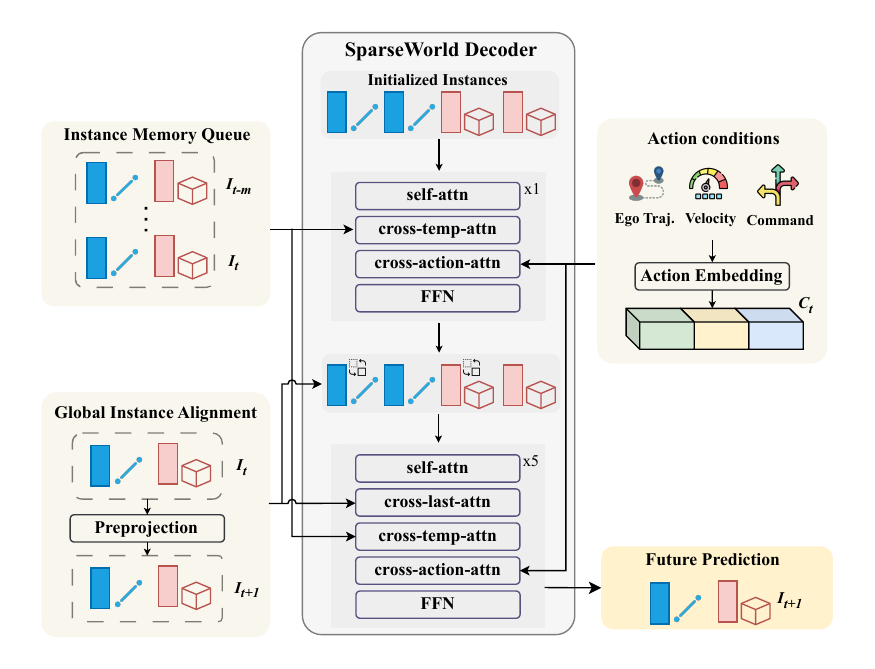}
      \captionsetup{font={small}}
      \caption{\textbf{Detailed structure of the Sparse Dreamer}, which predicts the subsequent frame’s instance in an autoregressive manner, based on historical instances and expected action conditions.}
      \label{fig:SparseDreamer}
      \vspace{-15pt} 
\end{figure}

\PAR{Sparse Dreamer.}
The module constitutes the most critical module for realizing future forecasting. As shown in Fig.~\ref{fig:SparseDreamer}, SPD leverages world modeling capabilities to predict future sparse instances, including agents and map elements. The SPD architecture comprises two components: (1) Global Instance Alignment, which processes one of $\mathcal{A}_{t} = \{a^\mathrm{agent}_{t}, a^\mathrm{map}_t\  \,|\, t \in \{0, 1, \cdots, f\}\}$ from IMQ, performing coarse pre-projection using trajectories and velocities predicted by FMP to enhance future forecasting. (2) SparseWorld Decoder, which employs an autoregressive transformer that ingests historical instances $\{I_{t-m}, \cdots, I_t\ \,|\, t \in \{0, 1, \cdots, f \}\}$, where $m$ 
denotes the length of historical instances selected from the IMQ,
and the current action condition $C_t$ from FMP to predict future instances $I_{t+1}$.

1) \textbf{Global Instance Alignment (GIA).}
Agent anchor $a^\mathrm{agent}$ contains 11-dimensional status information: center coordinates ($x,y,z$), dimensions ($w,l,h$), yaw representation ($\sin\psi,\cos\psi$), and velocity ($v_x,v_y,v_z$). To mitigate coordinate transformation errors in future frame prediction, GIA projects anchors to coarse future positions through ego-motion compensation and intrinsic velocity adjustment. For dynamic agents, we incorporate their own displacement:
\begin{equation}ce_{\mathrm{comp}}=ce-V\Delta t\end{equation}
where $V = (v_x, v_y, v_z)$ is the linear velocity of agent and $\Delta t$ is the time interval. $ce$ represents the center coordinates of the anchor box.
Using ego vehicle's angular velocity $\omega_t$ from $a^\mathrm{ego}$ and planner displacements ($\Delta x_t, \Delta y_t$) from the action condition $c_t$:
\begin{equation}
ce^{\prime} = R_z(\omega_t)ce_{\mathrm{comp}} + 
\left( \begin{array}{c}
\Delta x_t \\
\Delta y_t \\
0
\end{array} \right)
\end{equation}
\begin{equation}
\left( \begin{array}{c}
\cos\psi^{\prime} \\
\sin\psi^{\prime}
\end{array} \right) = R_z^{2\times2}(\omega_t)
\left( \begin{array}{c}
\cos\psi \\
\sin\psi
\end{array} \right)
\end{equation}
where $R_z$ is the Z-axis rotation matrix derived from angular velocity $\omega_t$. Static map elements ($V=0$) undergo ego-motion compensation only. This yields projected anchors $\mathcal{A}^{'}_{t}$ with updated spatial and kinematic status.

2) \textbf{SparseWorld Decoder (SWD).}
 SWD comprises $N$ transformer decoders, the initial decoder incorporates self-attention, temporal cross-attention, action-conditioned cross-attention, feed-forward networks, and classification and refinement modules. Subsequent decoders additionally integrate cross-attention with the outputs of preprojection modules. During temporal cross-attention operations, where historical instances $\{I_{t-m}, \cdots, I_t\ \,|\, t \in \{0, 1, \cdots, f \}\}$ originate from historical ego coordinate systems, we enhance spatial perception by injecting temporal embeddings $\mathrm{E_{time}} \in \mathbb{R}^{m\times c}$ and relative positional embeddings  $\mathrm{E_{pos}} \in \mathbb{R}^{m\times c}$, where $c$ represents the embedding dimensions. These positional embeddings are derived from motion planner trajectory proposals, which encode relative displacements ($\Delta x, \Delta y$) between historical frames and the current frame, and provide key spatial cues to mitigate coordinate transformation errors.

 To enhance the world model's environmental comprehension and controllable generation capabilities, we inject action conditions derived from the planner's outputs. 
 Following Drive-OccWorld~\cite{yang2025driving}, these action conditions are encoded as Fourier embeddings and integrated into the SparseWorld Decoder's attention mechanisms.

\subsection{Motion Planning Refinement}
\label{sec:Motion_and_Planning_Refine}
\PAR{Motion Prediction Refinement.}
Analogously to furnishing the motion predictor with a preliminary approximation of the solution, the world model predicts the future geometric information and features of agents and map elements, thereby providing a more robust reference for the motion predictor. 
Specifically, we reuse the motion predictor from the driving baseline, which takes current instance $I_0$ and historical instances $\{I_{-h},\dots,I_{-1}\}$ as input.
By substituting historical instances with future instances $\{I_1,\dots,I_f\}$, the predictor aggregates different spatial-temporal context with attention modules, resulting in refined motion trajectories with higher-quality and diverse motion patterns.
The refined motion predictions can serve either as superior alternatives to, or as effective supplements for, those from the driving baseline.

\begin{figure*}
    \centering
    \includegraphics[width=1.0\linewidth]{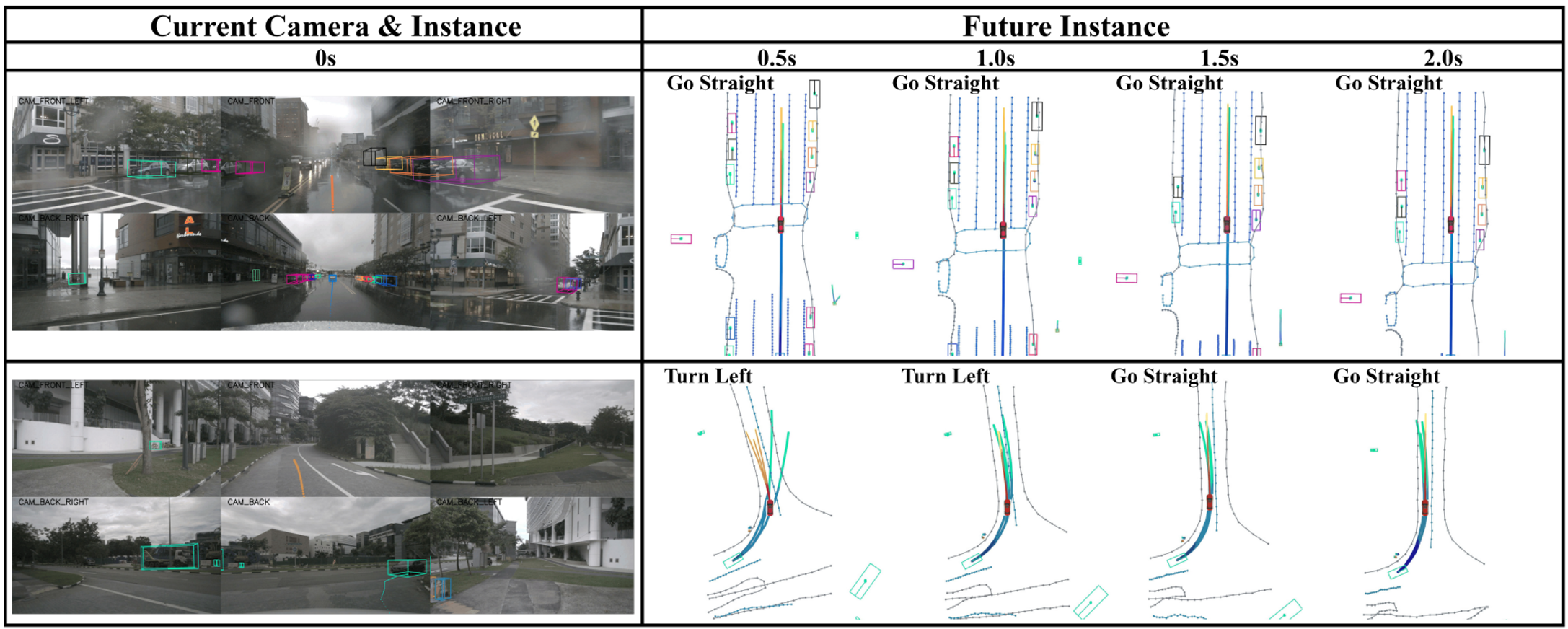}
    \caption{Qualitative results of future Instance forecasting. The results are presented at various future timestamps.}
    \label{fig:future_forecasting_vis}
    \vspace{-5pt} 
\end{figure*}
\begin{table*}[t]
\small
\caption{\textbf{Future Instance forecasting results on the nuScenes validation dataset.} \text{Avg.} denotes the average performance of that in 0.5s, 1.0s, 1.5s, and 2.0s. We use bold numbers to denote the best results.} 
\vspace{-1mm}
\centering 
\footnotesize
\setlength{\tabcolsep}{8.2pt} 
\newcolumntype{Y}{>{\centering\arraybackslash}X}
\begin{tabularx}{\linewidth}{l YYYYYY YYYYYY} 
\toprule 
& \multicolumn{6}{c}{Online Mapping/mAP $\uparrow$} & \multicolumn{6}{c}{3D Object Detection/NDS $\uparrow$} \\ 
\cmidrule(lr){2-7} \cmidrule(lr){8-13}
Method & $\text{0s}$ & $\text{0.5s}$ & $\text{1.0s}$ & $\text{1.5s}$ & $\text{2.0s}$ & $\text{Avg.}$ & $\text{0s}$ & $\text{0.5s}$ & $\text{1.0s}$ & $\text{1.5s}$ & $\text{2.0s}$ & $\text{Avg.}$\\ 
\midrule
copy\&paste & 56.66 & 33.23 & 19.18 & 12.81 & 9.71 & 18.73 & 0.523 & 0.331 & 0.285 & 0.267 & 0.212 & 0.274\\
projection & 56.66 & 48.61 & 38.03 & 27.49 & 21.84 & 33.99 & 0.523 & 0.459 & 0.433 & 0.412 & 0.390 & 0.424\\
\rowcolor{gray!20}
SparseWorld-S & 56.66 & \textbf{56.48} & \textbf{54.20} & \textbf{51.83} & \textbf{48.96} & \textbf{52.87} & 0.523 & \textbf{0.501} & \textbf{0.477} & \textbf{0.454} & \textbf{0.433} & \textbf{0.467}\\
\bottomrule
\end{tabularx}
\vspace{-5mm}
\label{table:Future_forecasting}
\end{table*}

\PAR{Trajectory Planning Refinement.}
Similar to motion prediction refinement, we first enhance the ego feature by interacting it with future instance features through attention modules, and then employ the same trajectory planning head as the driving baseline to generate the ego trajectories.
To explicitly reduce collision risk, we utilize distinct mechanisms for training and inference. In the training phase, a Safety-Critical Loss function is designed to guide the model toward low-risk trajectory generation. In the inference phase, an Adaptive Trajectory Selection module is applied to select and adjust the safest candidate trajectory in real time.

1) \textbf{Safety-Critical Loss (SCL).}
SCL extends the Ego-Agent Collision Constraint from VAD by introducing richer temporal and geometric modeling. Specifically, (1) SCL leverages future-frame agent positions and trajectories predicted by the world model—information that is inaccessible to VAD—and (2) explicitly models agent geometry, such as size and orientation, while the original VAD implementation only considers centroid coordinates.

Given an ego trajectory $\tau \in \mathbb{R}^{T \times 2}$ in the next $T$ steps and surrounding agent anchors $a^{agent}$ with their future trajectories $\tau^{agent}$,
safety-critical adjustment vector $v^{adj} \in \mathbb{R}^{T \times 2}$ is defined as the minimal perturbation necessary to ensure that $\tau$ satisfies safety constraints.
For each future step, we compute the bounding boxes of the ego vehicle and the agents based on their respective trajectories and geometric attributes. The required adjustments to the ego trajectory are then determined by the minimum distance vector between ego vehicle's and agent's bounding boxes and the safety distance threshold $\theta$.
The calculation process of $v^{adj}$ is denoted as:
\begin{equation}
{v}^{adj}=\text{SAV}(\tau,a^\mathrm{agent},\tau^{agent}, \theta)
\end{equation}
During training, the adjustment vector is derived from $a^{agent}$ and $\tau^{agent}$ of the baseline’s current frame, together with those predicted for future frames by FMP.
Denoting $v^{adj}_t$ as the item of $v^{adj}$ at step $t$,
the safety-critical loss $\mathcal{L}_{scl}$ is then computed as the average norm of the non-zero items of $v^{adj}$ across all steps. 
\begin{equation}
\mathcal{L}_{scl}=\frac{\sum_{t=1}^{T}||v^{adj}_t||_2}{\sum_{t=1}^{T}1_{\{||v^{adj}_i||_2 \neq0\}}}
\end{equation}

2) \textbf{Adaptive Trajectory Selection (ATS).}
During inference, we consider three candidate ego trajectories from various origins: the base trajectory predicted by the driving baseline, the future trajectory generated by the future forecasting module, and the refined trajectory optimized using the safety-critical loss. For each of these candidates, we perform collision detection and compute adjustment vectors and safety-critical losses based on both the baseline and refined motion prediction trajectories. The trajectory with the lowest safety-critical loss and no detected collisions is selected as the safest option for execution. In cases where all candidates are deemed equally safe, the refined trajectory is preferred. The final trajectory is obtained by applying the corresponding adjustment vector to the selected trajectory. More details of SCL and ATS are provided in the supplementary material.

\label{sec:experiments}
\section{Experiments}

\begin{table}[t!]
\small
\caption{
\textbf{Motion prediction results on the nuScenes validation dataset.} The Steps column indicates the length of the predicted future trajectories for evaluation.
}
\vspace{-1mm}
\centering 
\footnotesize
\setlength{\tabcolsep}{3.2pt}
\newcolumntype{Y}{>{\centering\arraybackslash}X}
\begin{tabularx}{\linewidth}{l Y YYYY} 
\toprule 
Method & Steps & $\text{EPA}\uparrow$ & $\text{MR}\downarrow$ & $\text{mADE}\downarrow$ & $\text{mFDE}\downarrow$\\ 
\midrule
UniAD~\cite{hu2023planning} & 12 & 0.456 & 0.151 & 0.71 & 1.02\\
SparseDrive-S~\cite{sun2024sparsedrive} & 12 & 0.482 & 0.136 & 0.62 & 0.99\\
\rowcolor{gray!20}
SparseWorld-S~(Ours) & 12 & \textbf{0.488} & \textbf{0.128} & \textbf{0.62} & \textbf{0.99}\\
\midrule
VAD-Tiny~\cite{jiang2023vad} & 6 & 0.598 & 0.121 & 0.79 & 1.08\\
\rowcolor{gray!20}
SparseWorld-V~(Ours) & 6 & \textbf{0.619} & \textbf{0.113} & \textbf{0.78} & \textbf{1.03}\\ 
\bottomrule
\end{tabularx}
\vspace{-5mm}
\label{table:motion_prediction}
\end{table}

\subsection{Experimental Setup}
\label{sec:experimental setup}

\PAR{Datasets}
To evaluate 3D detection, online mapping, motion prediction, and planning capabilities in real-world environments, we conduct extensive experiments on the nuScenes dataset~\cite{caesar2020nuscenes}. This dataset comprises 1,000 diverse driving scenes, each spanning approximately 20 seconds with 2Hz-annotated keyframes, partitioned into 700 training and 150 validation sequences. For closed-loop assessment of end-to-end planning performance, we employ the Bench2Drive~\cite{jia2024bench2drive} benchmark, which implements a CARLA Leaderboard v2.0-compatible evaluation protocol. Our evaluation follows official standards: training on the 1,000-clip base set and testing on 220 designated routes against established baselines. 
\begin{table}[t!]
\small
\caption{\textbf{Open-loop and Closed-loop Results of E2E Methods on Bench2Drive Benchmark.} Effi indicates Efficiency, Comf represents Comfortness, SR denotes Success Rate, DS refers to Driving Score.} 
\vspace{-1mm}
\centering 
\footnotesize
\setlength{\tabcolsep}{2.4pt}
\newcolumntype{Y}{>{\centering\arraybackslash}X}
\begin{tabularx}{\linewidth}{l cYYYY} 
\toprule 
& \multicolumn{1}{c}{\textbf{Open-loop}} & \multicolumn{4}{c}{\textbf{Closed-loop}} \\ 
\cmidrule(lr){2-2} \cmidrule(lr){3-6}
Method & $\text{Avg.L2$\downarrow$}$ & $\text{Effi$\uparrow$}$ & $\text{Comf$\uparrow$}$ & $\text{SR(\%)$\uparrow$}$ & $\text{\textbf{DS$\uparrow$}}$\\ 
\midrule
AD-MLP~\cite{zhai2023rethinking} & 3.64 & 48.45 & 22.63 & 0.00 & 18.05\\
UniAD-Tiny~\cite{hu2023planning} & 0.80 & 123.92 & 47.04 & 13.18 & 40.73\\
UniAD-Base~\cite{hu2023planning} & 0.73 & 129.21 & 43.58 & 16.36 & 45.81\\
SparseDrive-S~\cite{sun2024sparsedrive} & 0.87 & 170.21 & 48.63 & 16.71 & 44.54\\
MomAD~\cite{song2025don} & 0.82 & \textbf{174.91} & \textbf{50.98} & 18.11 & 47.91\\
\rowcolor{gray!20}
SparseWorld-S~(Ours) & 0.84 & 173.21 & 50.62 & \textbf{18.23} & \textbf{48.95}\\
\bottomrule
\end{tabularx}
\vspace{-5mm}
\label{table:closedloop_planning}
\end{table}

\begin{figure*}
    \centering
    \includegraphics[width=1.0\linewidth]{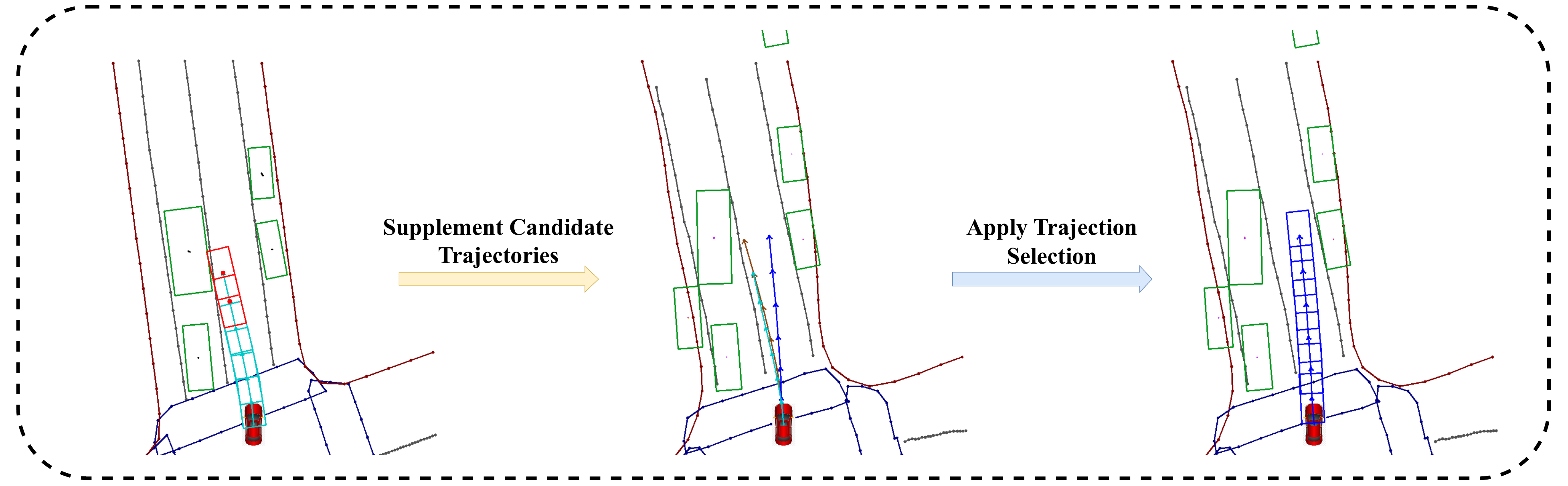}
    \caption{Qualitative results of adaptive trajectory selecting. In cases where the baseline trajectory results in inevitable collisions, the trajectories predicted by the world model and planning refiner can be leveraged as candidate alternatives to effectively avoid such collisions.}
    \label{fig:planning_vis}
    \vspace{-7pt} 
\end{figure*}

\begin{table*}[t!]
\small
\caption{\textbf{Planning results on the \textbf{nuScenes} validation dataset.} 
The results are derived from the evaluation protocol used in VAD~\cite{jiang2023vad}. 
We discard the ego status information to enable a fair comparison as shown in~\cite{li2024ego}.}
\vspace{-1mm}
\centering 
\footnotesize
\setlength{\tabcolsep}{10.6pt}
\newcolumntype{Y}{>{\centering\arraybackslash}X}
\begin{tabularx}{\linewidth}{l c c YYYY YYYY}
\toprule
& & & \multicolumn{4}{c}{L2 (m) $\downarrow$} & \multicolumn{4}{c}{Col. Rate (\%) $\downarrow$} \\
\cmidrule(lr){4-7} \cmidrule(lr){8-11}
Method & Input & Backbone & 1s & 2s & 3s & Avg. & 1s & 2s & 3s & Avg. \\
\midrule
UniAD~\cite{hu2023planning} & Camera & ResNet101 & 0.45 & 0.70 & 1.04 & 0.73 & 0.62 & 0.58 & 0.63 & 0.61 \\
Law~\cite{li2024enhancing} & Camera & Swin-T & \textbf{0.24} & \textbf{0.46} & \textbf{0.76} & \textbf{0.49} & 0.08 & 0.10 & 0.39 & 0.19 \\
MomAD~\cite{song2025don} & Camera & ResNet50 & 0.31 & 0.57 & 0.91 & 0.60 & 0.01 & 0.05 & 0.22 & 0.09 \\
\midrule
VAD-Tiny~\cite{jiang2023vad} & Camera & ResNet50 & 0.46 & 0.76 & 1.12 & 0.78 & 0.21 & 0.35 & 0.58 & 0.38 \\
\rowcolor{gray!20}
SparseWorld-V~(Ours) & Camera & ResNet50 & 0.32 & 0.56 & 0.88 & 0.59 & 0.08 & 0.18 & 0.46 & 0.24 \\ 
\midrule
SparseDrive-S~\cite{sun2024sparsedrive} & Camera & ResNet50 & 0.29 & 0.58 & 0.96 & 0.61 & 0.01 & 0.05 & 0.18 & 0.08 \\ 
\rowcolor{gray!20}
SparseWorld-S~(Ours) & Camera & ResNet50 & 0.33 & 0.62 & 1.01 & 0.65 & \textbf{0.00} & \textbf{0.01} & \textbf{0.13} & \textbf{0.05} \\ 
\bottomrule
\end{tabularx}
\vspace{-5mm}
\label{table:openloop_planning}
\end{table*}

\PAR{Baselines.}
To validate the flexibility and effectiveness of SparseWorld, we integrate it with two representative end-to-end autonomous driving frameworks, SparseDrive~\cite{sun2024sparsedrive} and VAD~\cite{jiang2023vad}. Both frameworks include instance perception modules for agents and maps, but differ in their perception paradigms. VAD is built upon a BEV-based dense perception paradigm, whereas SparseDrive adopts a sparse perception framework. In the following sections, we refer to the integrated versions as SparseWorld-S and SparseWorld-V, and present comprehensive evaluations to assess their performance. The experimental implementations are based on SparseDrive-S and VAD-Tiny, respectively.

\PAR{Tasks Definition.} 
We evaluate instance forecasting capabilities of SparseWorld and its enhancement of end-to-end driving system through three assessments: 
(1) \noindent\textbf{Agents and map layout forecasting} performance using established metrics, specifically nuScenes Detection Score (NDS) for detection and mean Average Precision (mAP) score for mapping. (2) \noindent\textbf{Motion prediction} refinement measured by motion metrics including minimum Average Displacement Error (minADE), minimum Final Displacement Error (minFDE), Miss Rate (MR), and End-to-End Prediction Accuracy (EPA). (3) \noindent\textbf{Trajectory planning} performance evaluated through open-loop metrics such as L2 Displacement Error and Collision Rate, with closed-loop assessment using comprehensive metrics from Bench2Drive including Driving Score (DS), Success Rate (SR), Comfortness (Comf), and Efficiency (Effi).

\subsection{Main Results}
\label{sec:main results}
\PAR{Agents and Map Layout Forecasting Results.}
Table~\ref{table:Future_forecasting} illustrates the online mapping and 3D detection results in the future. The number at 0s represents the perception metircs of SparseDrive-S~\cite{sun2024sparsedrive}, while 0.5s-2s measure performance of SparseWorld forecasting. Two reference methods are compared: (1) Copy\&Paste directly reuses 0s perception at future timestamps. (2) Projection transforms 0s detections using ground truth projection matrices. Results demonstrate that SparseWorld achieves \noindent\textbf{82.8\%} (3D detection) and \noindent\textbf{86.4\%} (online mapping) of the baseline performance at 2s, substantially outperforming both reference methods by significant margins, indicating successful learning of scene evolution dynamics within our model.
Additionally, in Fig.~\ref{fig:future_forecasting_vis}, we demonstrate the capability of SparseWorld to forecast future instances from current frame representations across diverse scenarios, validating the potential of SparseWorld to generate distinct plausible future instances conditioned on varying control commands.

\PAR{Motion prediction Results.}
Table~\ref{table:motion_prediction} presents the performance of SparseWorld on the motion prediction task. 
SparseWorld consistently enhances the motion predictor across both architectures. When coupled with SparseDrive-S, SparseWorld-S achieves a superior EPA of 0.488 and notably reduces the MR to 0.128, demonstrating more comprehensive motion patterns while maintaining the baseline's strong mADE/mFDE scores. For the VAD-Tiny baseline, SparseWorld-V yields holistic improvements across all evaluation metrics, elevating the EPA to 0.619 and reducing the MR to 0.113. These results highlight the effectiveness of our motion refinement strategy.

\PAR{Open-loop Planning Results.}
As illustrated in Table~\ref{table:openloop_planning}, SparseWorld offers a comprehensive enhancement to the mid-tier VAD-Tiny architecture, effectively reducing the average L2 error from $0.78\text{m}$ to $0.59\text{m}$ and the collision rate from $0.38\%$ to $0.24\%$. 
Furthermore, on the high-performing SparseDrive-S baseline, SparseWorld-S demonstrates superior efficacy, successfully optimizing the average collision rate from $0.08\%$ to a minimal $0.05\%$, achieving SOTA performance. 
Table~\ref{fig:planning_vis} provides a visual illustration of the effectiveness of the trajectory the Adaptive Trajectory Selection module. Additional experimental results are provided in the supplementary material.

\PAR{Closed-loop Planning Results.}
We present challenging closed-loop evaluation results on the Bench2Drive dataset, as detailed in Table~\ref{table:closedloop_planning}. 
The assessment covers 220 routes, with each route containing a safety-critical scenario representing 44 distinct interaction types. 
SparseWorld-S demonstrates significant improvements over SparseDrive, achieving a \noindent\textbf{9.9\%} increase respectively in the key Driving Score metric, while also improving the Success Rate metric by \noindent\textbf{9.1\%}. 
These results validate SparseWorld's effectiveness as an enhancement module for boosting closed-loop performance in end-to-end driving systems.

\begin{table}[t]
\small
\caption{\textbf{Memory usage and inference time tested across different generative levels.} IM refers to Inference Memory, IPT indicates Inference Perception Time, IGT represents Inference Generation Time.} 
\vspace{-1mm}
\centering 
\footnotesize
\setlength{\tabcolsep}{3pt}
\newcolumntype{Y}{>{\centering\arraybackslash}X}
\begin{tabularx}{\linewidth}{l cYYY} 
\toprule 
Method & $\text{Representation}$ & $\text{IM}$ & $\text{IPT}$ & $\text{IGT}$\\ 
\midrule
$\text{Uniscene}$~\cite{li2025uniscene} & Image & 23524$\text{M}$ & $\text{-}$ & 28.29$\text{s}$\\
$\text{Drive-OccWorld}$~\cite{yang2025driving} & Occupancy & 20581$\text{M}$ & 401$\text{ms}$ & 398$\text{ms}$\\
\midrule
\rowcolor{gray!20}
$\text{SparseWorld-S~(Ours)}$ & Instance & \textbf{4397$\text{M}$} & \textbf{174$\text{ms}$} & \textbf{70$\text{ms}$}\\
\bottomrule
\end{tabularx}
\vspace{-3mm}
\label{table:efficiency}
\end{table}

\begin{table}[t]
\small
\caption{\textbf{Collision Rate Gains Across Different World Models.} $\text{Drive-OccWorld}^{\text{-wm}}$ denotes the variant without the world model. }
\vspace{-1mm}
\centering 
\footnotesize
\setlength{\tabcolsep}{3pt}
\newcolumntype{Y}{>{\centering\arraybackslash}X}
\begin{tabularx}{\linewidth}{l cYYYc} 
\toprule 
& & \multicolumn{4}{c}{Col. Rate (\%) $\downarrow$} \\
\cmidrule(lr){3-6}
Method & $\text{Representation}$ & $\text{0.5s}$ & $\text{1.0s}$ & $\text{1.5s}$ & $\text{Avg.}$\\ 
\midrule
$\text{Drive-OccWorld}^{\text{-wm}}$~\cite{yang2025driving} & Occupancy & 0.16 & 0.23 & 0.32 & 0.24\\
$\text{Drive-OccWorld}$~\cite{yang2025driving} & Occupancy & 0.04 & 0.11 & 0.13 & 0.09\\
\midrule
$\text{VAD-Tiny}$~\cite{jiang2023vad} & Instance & 0.15 & 0.21 & 0.29 & 0.22\\
\rowcolor{gray!20}
$\text{SparseWorld-V~(Ours)}$ & Instance & \textbf{0.05} & \textbf{0.08} & \textbf{0.12} & \textbf{0.08}\\
\midrule
$\text{SparseDrive-S}$~\cite{sun2024sparsedrive} & Instance & 0.01 & 0.02 & 0.03 & 0.02\\
\rowcolor{gray!20}
$\text{SparseWorld-S~(Ours)}$ & Instance & \textbf{0.00} & \textbf{0.00} & \textbf{0.00} & \textbf{0.00}\\
\bottomrule
\end{tabularx}
\vspace{-5mm}
\label{table:effectiveness}
\end{table}

\PAR{Comparison with Other World Models.} 
SparseWorld demonstrates clear advantages in both efficiency and effectiveness when benchmarked against contemporary world models, as summarized in Table~\ref{table:efficiency} and Table~\ref{table:effectiveness}.
Regarding efficiency, we compare our instance-level generation (900 agent anchors, 100 map anchors) with UniScene (image generation, $512 \times 1536$) and $\text{Drive-OccWorld}$ (occupancy prediction, $200 \times 200 \times 16$). Notably, our approach generates 4-frame instance-level scenes in just 70 ms using only 4 GB of memory, offering a substantial efficiency gain over both alternatives.
In terms of effectiveness (Table~\ref{table:effectiveness}), the collision mitigation achieved by our sparse representation clearly surpasses $\text{Drive-OccWorld}$’s dense paradigm. 
While introducing WM reduced $\text{Drive-OccWorld}$’s 1.5 s collision rate by 62.5\%, our $\text{SparseWorld}$ achieved even greater gains—lowering $\text{VAD-Tiny}$’s collision rate by 63.6\% and reducing $\text{SparseDrive-S}$’s to 0\%. This marked disparity indicates that our sparse world model is inherently more effective for downstream planning, validating its superiority over dense representations.

\begin{table}[t!]
\small
\caption{Abalation on Sparse Dreamer.} 
\vspace{-1mm}
\centering 
\footnotesize
\setlength{\tabcolsep}{8.4pt}
\newcolumntype{Y}{>{\centering\arraybackslash}X}
\begin{tabularx}{\linewidth}{YY YYYYY} 
\toprule 
\multicolumn{7}{c}{Online Mapping/mAP $\uparrow$}\\ 
\midrule
PE & PP & $\text{0.5s}$ & $\text{1.0s}$ & $\text{1.5s}$ & $\text{2.0s}$ & $\text{Avg.}$\\ 
\midrule
{$\bm \checkmark$} & {$\bm \times$} & 55.28 & 53.16 & 51.07 & 47.76 & 51.82\\
{$\bm \times$} & {$\bm \checkmark$} & 55.11 & 53.36 & 51.30 & 48.54 & 52.08\\
{$\bm \checkmark$} & {$\bm \checkmark$} & \textbf{56.48} & \textbf{54.20} & \textbf{51.83} & \textbf{48.96} & \textbf{52.87}\\
\midrule 
\multicolumn{7}{c}{3D Object Detection/NDS $\uparrow$}\\ 
\midrule
PE & PP & $\text{0.5s}$ & $\text{1.0s}$ & $\text{1.5s}$ & $\text{2.0s}$ & $\text{Avg.}$\\ 
\midrule
{$\bm \checkmark$} & {$\bm \times$} & 49.19 & 46.68 & 44.27 & 42.10 & 45.54\\ 
{$\bm \times$} & {$\bm \checkmark$} & 49.26 & 46.85 & 44.64 & 42.56 & 45.83\\
{$\bm \checkmark$} & {$\bm \checkmark$} & \textbf{50.13} & \textbf{47.73} & \textbf{45.44} & \textbf{43.33} & \textbf{46.66}\\
\bottomrule
\end{tabularx}
\vspace{-3mm}
\label{table:abalation_future_forecasting}
\end{table}

\begin{table}[t!]
\small
\caption{Abalation on Motion Prediction.} 
\vspace{-1mm}
\centering 
\footnotesize
\setlength{\tabcolsep}{4.7pt}
\newcolumntype{Y}{>{\centering\arraybackslash}X}
\begin{tabularx}{\linewidth}{YYYYYYY} 
\toprule 
Agent & Map & Steps & $\text{EPA}\uparrow$ & $\text{MR}\downarrow$ & $\text{mADE}\downarrow$ & $\text{mFDE}\downarrow$\\ 
\midrule
$\bm \times$ & $\bm \times$ & 6 & 0.598 & 0.121 & 0.79 & 1.08\\
$\bm \checkmark$ & $\bm \times$ & 6 & 0.610 & 0.116 & 0.79 & 1.06\\
$\bm \times$ & $\bm \checkmark$ & 6 & 0.612 & 0.115 & 0.79 & 1.05\\
$\bm \checkmark$ & $\bm \checkmark$ & 6 & \textbf{0.619} & \textbf{0.113} & \textbf{0.78} & \textbf{1.03}\\
\bottomrule
\end{tabularx}
\vspace{-3mm}
\label{table:ablation_motion_prediction}
\end{table}

\begin{table}[t!]
\caption{Ablation on Trejectory Planning.} 
\vspace{-1mm}
\centering 
\footnotesize
\setlength{\tabcolsep}{7.2pt}
\newcolumntype{Y}{>{\centering\arraybackslash}X}
\begin{tabularx}{\linewidth}{YYY YYYY YYYY} 
\toprule 
& & &  \multicolumn{4}{c@{}}{L2(\text{m}) $\downarrow$} & \multicolumn{4}{c@{}}{Col. Rate(\%) $\downarrow$}\\ 
\cmidrule(lr){4-7} \cmidrule(lr){8-11}
FIF & SCL & ATS & $\text{1s}$ & $\text{2s}$ & $\text{3s}$ & $\text{Avg.}$ & $\text{1s}$ & $\text{2s}$ & $\text{3s}$ & $\text{Avg.}$\\ 
\midrule
$\bm \times$ & $\bm \times$ & $\bm \times$ & 0.46 & 0.76 & 1.12 & 0.78 & 0.21 & 0.35 & 0.58 & 0.38\\
$\bm \checkmark$ & $\bm \times$ & $\bm \times$ & 0.35 & 0.60 & 0.92 & 0.62 & 0.16 & 0.31 & 0.56 & 0.34\\
$\bm \checkmark$ & $\bm \checkmark$ & $\bm \times$ & 0.32 & 0.57 & 0.88 & 0.59 & 0.10 & 0.23 & 0.50 & 0.28\\
$\bm \checkmark$ & $\bm \checkmark$ & $\bm \checkmark$ & \textbf{0.32} & \textbf{0.56} & \textbf{0.88} & \textbf{0.59} & \textbf{0.08} & \textbf{0.18} & \textbf{0.46} & \textbf{0.24}\\
\bottomrule
\end{tabularx}
\vspace{-5mm}
\label{table:ablation_planning}
\end{table}

\subsection{Ablation Study}
\PAR{Ablation on Sparse Dreamer.}
We base this ablation study on the $\text{SparseWorld-S}$. As shown in Table~\ref{table:abalation_future_forecasting}, our ablation study on future prediction tasks investigates two feature enhancement strategies: future pre-projection (PP) in the Global Instance Alignment and relative positional embeddings (PE) of historical frames in the SparseWorld Decoder. The results demonstrate that both enhancements contribute positively to performance, with future pre-projection delivering particularly significant gains. 

\PAR{Ablation on Motion Prediction.}
We conduct ablation studies on \text{SparseWorld-V} to investigate the contribution of different types of future instances to our motion refinement strategy, as shown in Table~\ref{table:ablation_motion_prediction}. The results demonstrate the effectiveness of both Agent and Map instances, each providing notable individual improvements. When combined, these components exhibit a synergistic effect, yielding the most significant overall gains and achieving superior performance across all metrics. 

\PAR{Ablation on trajectory planning.}
Table ~\ref{table:ablation_planning} summarizes the influence of each component within the trajectory optimization strategy on overall planning performance. In \text{SparseWorld-V}, integrating future instance features (FIF) primarily contributes to reducing the L2 error, whereas the introduction of the Safety-Critical Loss (SCL) markedly decreases the collision rate. Moreover, the Adaptive Trajectory Selection (ATS) further suppresses collisions while maintaining comparable L2 accuracy, showing its effectiveness in enhancing safety without sacrificing precision.


\section{Conclusion}
\label{sec:conclusion}
We propose SparseWorld, a lightweight and efficient world model explicitly tailored for instance-level future forecasting and motion planning refinement. It first predicts high-fidelity future sparse scene representations, which are subsequently utilized to enhance both motion prediction and trajectory planning. Extensive experimental results consistently demonstrate that the model’s accurate future forecasting capability effectively informs end-to-end autonomous driving, leading to substantial improvements in both safety and robustness.

\begingroup
\AtNextBibliography{\footnotesize}
\printbibliography
\endgroup

\end{document}